\documentclass[journal]{IEEEtran}

\usepackage{times}
\usepackage{epsfig}
\usepackage{graphicx}
\usepackage{amsmath}
\usepackage{amssymb}

\usepackage{subfigure}

\usepackage{algorithm}
\usepackage{algorithmicx}
\usepackage{algpseudocode}
\usepackage{booktabs}
\usepackage{multirow}
\usepackage{color}
\usepackage{hyperref}

\usepackage{url}
\ifCLASSINFOpdf
\else
\fi
\begin{document}

%
\title{Object Detection based on OcSaFPN in Aerial Images with Noise}
%
%
%

\author{~Chengyuan~Li,~Jun~Liu,~Hailong~Hong,~Wenju~Mao,~Chenjie~Wang,~Chudi~Hu,~Xin~Su,~Bin~Luo}

\maketitle

\begin{abstract}
Taking the deep learning-based algorithms into account has become a crucial way to boost object detection performance in aerial images. While various neural network representations have been developed, previous works are still inefficient to investigate the noise-resilient performance, especially on aerial images with noise taken by the cameras with telephoto lenses, and most of the research is concentrated in the field of denoising. Of course, denoising usually requires an additional computational burden to obtain higher quality images, while noise-resilient is more of a description of the robustness of the network itself to different noises, which is an attribute of the algorithm itself. For this reason, the work will be started by analyzing the noise-resilient performance of the neural network, and then propose two hypotheses to build a noise-resilient structure. Based on these hypotheses, we compare the noise-resilient ability of the Oct-ResNet with frequency division processing and the commonly used ResNet. In addition, previous feature pyramid networks used for aerial object detection tasks are not specifically designed for the frequency division feature maps of the Oct-ResNet, and they usually lack attention to bridging the semantic gap between diverse feature maps from different depths. On the basis of this, a novel octave convolution-based semantic attention feature pyramid network (OcSaFPN) is proposed to get higher accuracy in object detection with noise. The proposed algorithm tested on three datasets demonstrates that the proposed OcSaFPN achieves a state-of-the-art detection performance with Gaussian noise or multiplicative noise. In addition, more experiments have proved that the OcSaFPN structure can be easily added to existing algorithms, and the noise-resilient ability can be effectively improved.
\end{abstract}


%
\IEEEpeerreviewmaketitle

\section{Introduction}
\label{sec:Introduction}
%
%
%
%
\IEEEPARstart{D}{etecting} objects in aerial images is a fundamental but challenging task of image processing \cite{ding2019learning}. With the rapid development of sensor technologies and neural networks, new applications and demands based on aerial imagery have emerged, such as traffic control, monitoring of oil storage facilities, urban village inspection and military target discovery \cite{2019Sig, 2020Object}. Although many methods have been proposed, object detection in aerial images with noise is still pretty challenging.

Widely used convolutional neural networks (CNNs) greatly promoted the development of object detection technology. A region-based CNN (R-CNN) \cite{2014Rich}, deploys an algorithm called selective search \cite{2013Selective} to predict the possible regions of interest, and then uses a support vector machine (SVM) for classification and localization. R-CNN shows the great potential of CNNs in feature description, but this method does not make full use of multi-scale feature maps, and the performance of detection with drastic scale changes is not robust. More importantly, R-CNN is not an end-to-end network with slow detection speed. To address this issue, SPP-Net \cite{2014Spatial} extracts features from the whole image once, and then shares the features during the detection, avoiding the inefficiency of repeated feature extraction. Besides, Fast R-CNN \cite{girshick2015fast} and Faster R-CNN \cite{Ren2017Faster} use neural network structures to replace the SVM and the selective search respectively to achieve fast end-to-end object detection. In addition, the feature pyramid network (FPN) \cite{2017Feature} makes full use of both the feature maps with different depths, making the algorithm more robust in dealing with objects with large scale changes. The combination of Faster R-CNN and FPN forms an end-to-end two-stage detector, which can meet the high-precision object detection task. As a result, many newer networks have designed on this baseline \cite{2017Mask, li2017light, cai2018cascade, lin2017focal, kong2016hypernet, dai2017deformable, jiang2017r2cnn, ma2018arbitrary}.

Much researches on optical aerial image object detection focus on the two-stage detectors, due to the high precision. According to the type of the bounding box, these algorithms can be divided into two categories, regular rectangular bounding box and rotated rectangular bounding box. Regular rectangular bounding box-based detectors appear earlier. Zou et al. \cite{zou2016ship} designed a robust method, called singular value decomposition networks (SVDNet), for ship detection tasks which usually interfered with by clouds, strong waves, and high computational expenses. The SVDNet combines singular value decomposition with a convolutional neural network to achieve high-precision aerial object detection, but the method of extracting candidate regions of this algorithm is not efficient enough. Sig-NMS \cite{dong2019sig} optimized the non-maximum suppression (NMS), which improves the ability of networks to detect small objects. Both Deng et al. \cite{deng2017toward} and Xiao et al. \cite{xiao2017airport} have used sliding windows algorithm to extract region proposals, and then use CNNs to extract features to achieve object detection in aerial images. In fact, the sliding windows algorithm is relatively inefficient and can be replaced by a sophisticated neural network structure. Furthermore, Xu et al. \cite{xu2017deformable} and Ren et al. \cite{ren2018deformable} used deformable convolution \cite{dai2017deformable} in remote sensing image object detection. Similar to the dilated convolution \cite{yu2015multi}, deformable convolution changes the receptive field of the convolutional kernel and introduces more semantically related information. As for rotated rectangular bounding box-based detectors, these algorithms believe that the use of a more accurate bounding box representation method can better describe the object under the overhead view. Similar to the rotation region proposal networks (RRPN) \cite{ma2018arbitrary}, both Li et al. \cite{li2017rotation} and RoITransformer \cite{ding2019learning} obtained more accurate rotated region proposals in the region proposal networks stage, and minimize the number of background pixels in the region proposals to improve detection accuracy. Yang et al. \cite{yang2019scrdet} constructed a remote sensing image object detector called SCRDet for predicting rotate rectangular coordinates using an attention mechanism. CAD-Net \cite{zhang2019cad} and Li et al. \cite{2020Object} tried to use attention-based branches to capture the potential connection between the scene and the object.

An alternative of two-stage deep learning methods is one-stage-based regression that uses a single CNN to predict bounding boxes and class probabilities directly from full images in one evaluation. As for natural image object detection tasks, one-stage detectors have the advantage of fast speed, but their accuracy is often slightly inferior to the two-stage detectors. Typical algorithms include you only look once (YOLO) \cite{redmon2016you}, single shot multibox detector (SSD) \cite{liu2016ssd}, CornerNet \cite{law2018cornernet} and FCOS \cite{tian2019fcos}, etc. Remote sensing images have complex backgrounds, overhead viewing angles, small target problem, imbalance in the number of foreground and background areas, etc., resulting in less application of one-stage detectors in remote sensing image object detection. You only look twice (YOLT) \cite{van2018you} tried to detect objects and scenes at the same time. FMSSD \cite{wang2019fmssd} used an atrous spatial feature pyramid on the basis of the SSD to improve the robustness to the scale changes on remote sensing images. Zou et al. \cite{zou2017random} proposed a small object detection method based on Bayesian prior conditions in high-resolution aerial images. R3Det \cite{yang2019r3det} designed a feature refinement module (FRM) to get rotated bounding box-based results.

\begin{figure}[t]
	\begin{center}
		\includegraphics[width=1.0\linewidth]{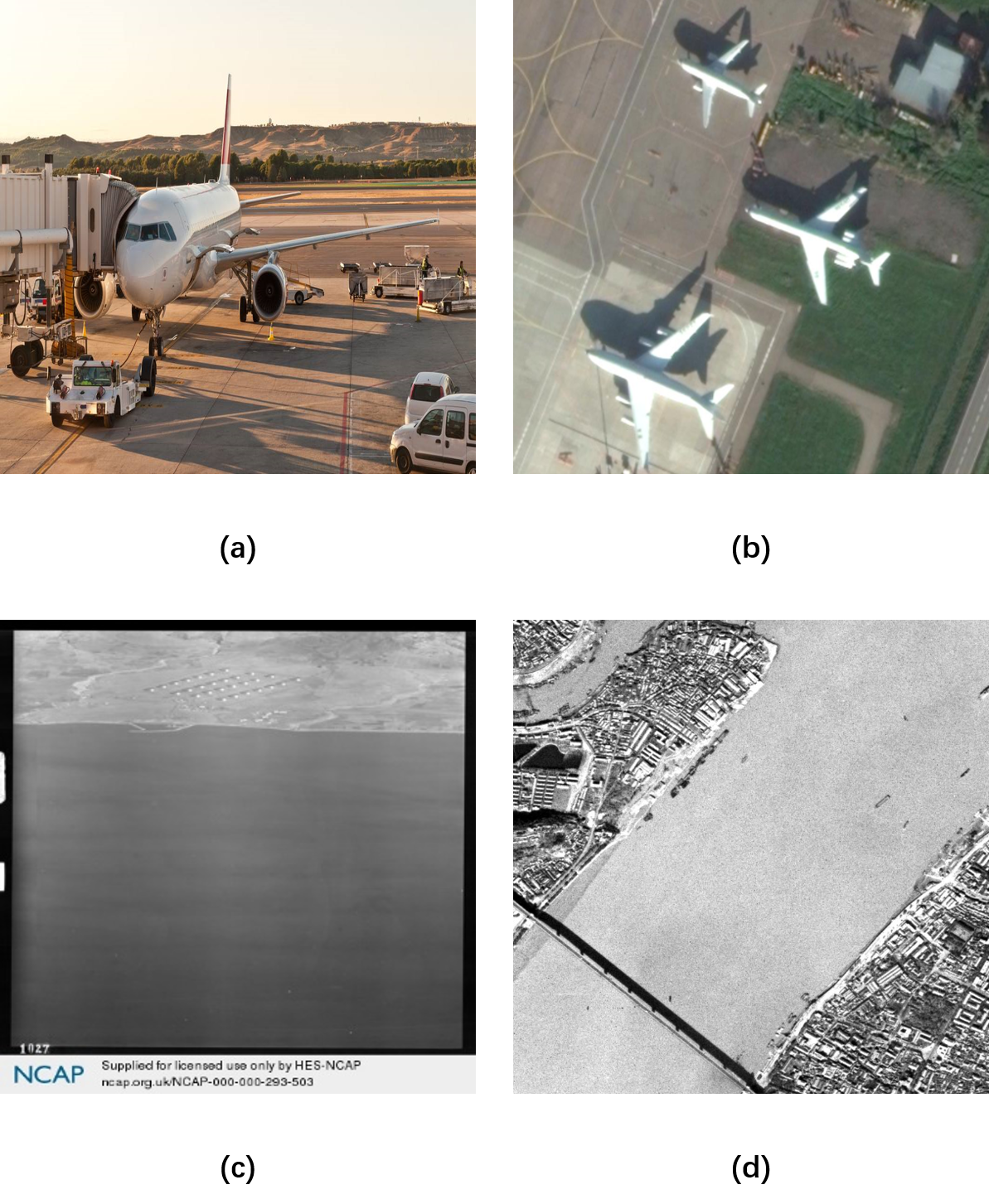}
	\end{center}
	\caption{Different types of images: (\textbf{a}) A natural image from a surveillance camera. (\textbf{b}) An aerial image from DOTA dataset \cite{Xia2018DOTA}. Unlike natural images, remote sensing images usually have a top-view perspective and have a complex background. (\textbf{c}) A reconnaissance image taken during World War II. This kind of remote sensing image is usually taken with a squint angle of view, and the noise is serious due to the long atmosphere. (\textbf{d}) A typical intelligence satellite image from the U.S. Geological Survey (USGS) \cite{2013Declassified}.}
	\label{fig:fig1}
\end{figure}

Although many algorithms have focused on the problem of rotation and multi-scale detection in remote sensing image object detection, however, there remains a primary challenge in optical remote sensing images with noise. In fact, aerial images with noise are very common, and they usually come from aerial platforms such as special cameras with telephoto lenses on long-range unmanned aerial vehicles (UAVs) or old satellites (e.g. declassified intelligence satellite photography, DISP). As shown in Fig.~\ref{fig:fig1}(c), for images from cameras with telephoto lenses on long-range UAVs, during imaging, there are often serious atmospheric noises due to the long atmospheric distance through. Therefore, the generated images usually have characteristics of low signal-to-noise ratio, high spatial resolution. This type of image is often used in military or mapping fields \cite{huang2007calibrating,tarasov2016research,fujiwara2019super}. As shown in Fig.~\ref{fig:fig1}(d), images from DISP missions, also have a lot of noise. This kind of images can provide a lot of historical information and is widely used in historical investigation, change detection, and other fields \cite{casana2020global,rendenieks2020half,tian2019applications}. For the above types of data, high-frequency information loss makes it difficult to recover high-quality images. The low-frequency information is filled with a large amount of noise, which makes detecting useful objects need extra efforts other than normal object detection tasks on such low-quality images. Thus, detecting aerial images with noise needs extra efforts other than normal object detection approaches.

\begin{figure}[t]
	\begin{center}
		\includegraphics[width=1.0\linewidth]{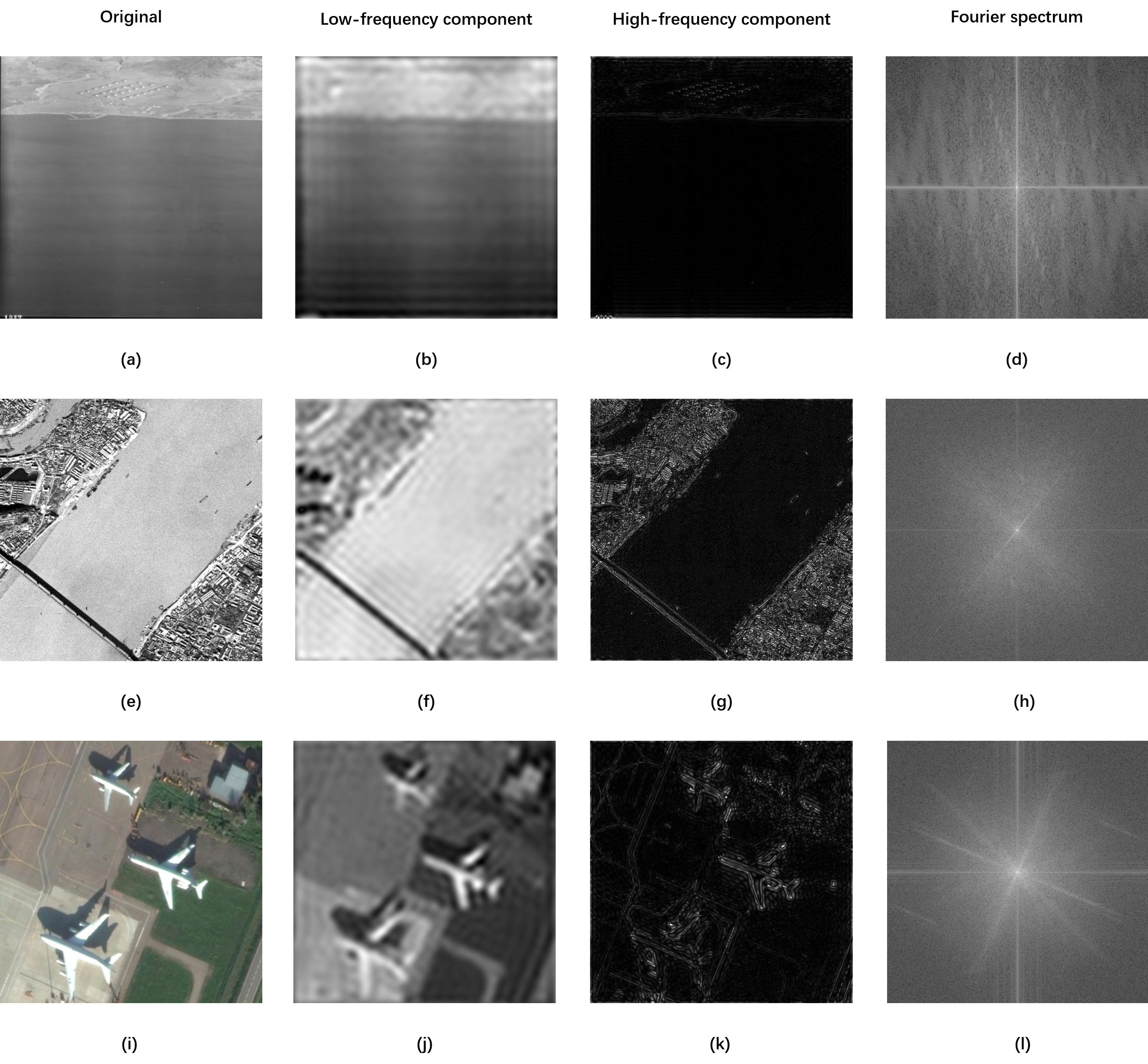}
	\end{center}
	\caption{Visualization of low-frequency components, high-frequency components, and Fourier spectrums of three different images. From left to right, each line is: original image, low-frequency component, high-frequency component, and Fourier spectrum. The image of the first line is a reconnaissance image taken during World War II (images from DISP missions with noise), the image of the second line is a typical intelligence satellite image from the USGS (images from DISP missions with noise), and the image of the third line is a common aerial image from DOTA dataset.}
	\label{fig:fig2}
\end{figure}

The basic motivation of this paper is to enable the neural network itself to get stronger noise-resilient capability without more computational requirements like common denoising algorithms, so that the algorithm can have better robustness under different noise situations as possible. In fact, there are few researches on the noise-resilient performance of different neural network structures at present. Inspired by the Fourier transformation \cite{goda2013dispersive} and the wavelet transformation \cite{akansu1996multiplierless}, we try to understand the neural network from the frequency domain. As shown in Fig.~\ref{fig:fig2}, compared to the common aerial image (Fig.~\ref{fig:fig2}(i)), images from DISP missions (Fig.~\ref{fig:fig2}(a) and (e)) always have more noise in their low-frequency components, and they also lose more useful information in low-frequency components. Therefore, for the deep learning-based object detection algorithms, we propose a hypothesis that the frequency division processing structure such as octave convolution \cite{chen2019drop} can improve the noise-resilient capability of the network. Based on this hypothesis, a novel octave convolution-based semantic attention feature pyramid network (OcSaFPN) is proposed. To be specific, OcSaFPN construction employs the frequency-divided feature map as input and then makes the learned convolution kernel smoother through the message passing operation similar to octave convolution. Besides, a novel multi-scale feature fusion strategy is applied to OcSaFPN to reduce the semantic gap between feature maps of different depths. The unified framework achieves state-of-the-art performance on two public large-scale datasets and one dataset based on the DOTA dataset after processing. Meanwhile, more experiments have proved that OcSaFPN can plug and play, rapidly improving the noise-resilient capability of the existing high-precision algorithms. The main contributions of this article are as follows:

\begin{itemize}
  \item [1)]
  To our knowledge, this is the first work that considers the noise-resilient performance of the neural network structure itself in the remote sensing image object detection. We specifically analyze the DISP data, then study the noise-resilient performance of the neural network from the perspective of the frequency domain, and propose a hypothesis that compressing the low frequency and reducing the redundant information can improve the noise-resilient ability of the network. Based on this hypothesis, the Oct-ResNet \cite{chen2019drop} and the ResNet \cite{he2016deep} are compared and tested to verify the effect of frequency-division convolution on noise-resilient abilities.
  \item [2)]
  To better fit the backbone structure of the frequency division processing, we design an OcSaFPN module which contains the similar message passing operation to the octave convolution to get noise-resilient capability and a feature fusion module to unify the context information from different feature maps.
  \item [3)]
  A large number of experiments demonstrate the effectiveness of OcSaFPN structure proposed in this paper. We evaluate OcSaFPN equipped with several detectors on three datasets and it consistently brings significant improvements over FPN based detectors.
\end{itemize}

The rest of this paper is organized as follows. Section~\ref{sec:RelatedWork} gives a brief review of the related work on octave convolution, image processing techniques for noise images, and the multi-scale feature representations. In Section~\ref{sec:ProposedMethod}, we introduce the proposed method in detail. The details of the datasets, the experiments conducted in this study, and the results of the experiments are presented in Section~\ref{sec:Experimental}. Conclusions are drawn in Section~\ref{sec:Conclusion}.

\section{Releated Work}
\label{sec:RelatedWork}

\subsection{Octave Convolution}
\label{sec:OctaveConvolution}

\begin{figure}[t]
	\begin{center}
		\includegraphics[width=1.0\linewidth]{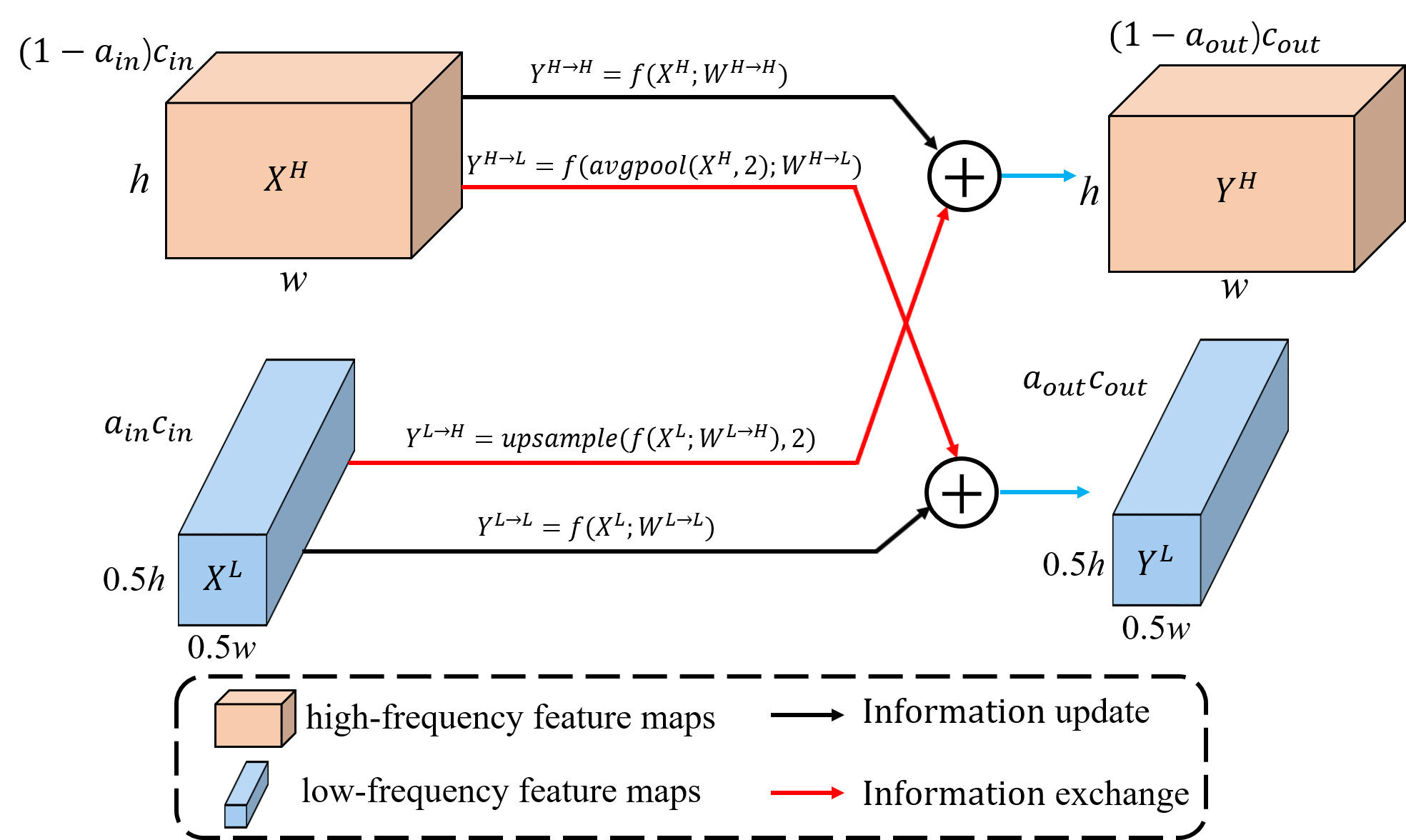}
	\end{center}
	\caption{Detailed design of the Octave Convolution. Black arrows correspond to information updates while red arrows facilitate information exchange between the two frequencies \cite{chen2019drop}.}
	\label{fig:octave}
\end{figure}

The original approach towards octave convolution has been introduced by Chen et al. \cite{chen2019drop}. Unlike the ordinary convolution, the octave convolution treats the features in ordinary convolution as a combination of high-frequency components and low-frequency components (see Fig.~\ref{fig:octave} for details). Formally, a input feature tensor can be seen as $X\in\mathbb{R}^{c\times h\times w}$, where $h$, $w$ and $c$ denote the height, the width and the number of channels. With a ratio $\alpha\in[0,1]$, the $X$ can be represented as $X=\{X^H,X^L\}$, where the high-frequency feature maps $X^H\in{\mathbb{R}}^{(1-\alpha)c\times h \times w}$ capture fine details and the low-frequency maps $X^L\in{\mathbb{R}}^{\alpha c \times \frac{h}{2} \times \frac{w}{2}}$ vary slower in the spatial dimensions. Based on this feature map representation, the convolution process can be expressed by the following formulas:

\begin{equation}
\begin{split}
Y^H &= f(X^H;W^{H\rightarrow H}) + upsample(f(X^L;W^{L\rightarrow H}),2) \\
Y^L &= f(X^L;W^{L\rightarrow L}) + f(avgpool(X^H,2);W^{H\rightarrow L}),
\label{eq:octave}
\end{split}
\end{equation}

where $X,Y$ denote the input tensor and the output tensor, $f(X;W)$ ia a convolution with parameters $W$, $avgpool(X,k)$ is an average pooling operation with kernel size $k\times k$ and stride $k$. The symbol of $upsample(X,k)$ is an up-sampling operation by a factor of $k$ via nearest interpolation.

The motivation of the octave convolution is to improve network efficiency and reduce the demand of hardware resources by reducing the spatially redundancy of feature map. In fact, some attempts have been made before octave convolution. Ke et al. \cite{ke2017multigrid} designed a multigrid extension of convolutional neural networks. With within-scale and cross-scale extent, the proposed method can achieve the internal information exchange and dynamic routing mechanism between multi-scale features. Chen et al. \cite{article} designed a wavelet-like auto-encoder structure to divide the normal feature map into high-frequency and low-frequency maps with L2 norm. Briefly, since the energy of the image is more concentrated in the low-frequency components, the corresponding high-frequency components can be obtained by minimizing the L2 norm. At present, many works have been done to improve network performance based on octave convolution. Xu et al. \cite{xu2020csa} and Tang et al. \cite{tang2020hyperspectral}  focus on the applications of 3-D octave convolution. Ricard et al. \cite{DBLP:journals/corr/abs-1905-12534} proposed a variety of octave convolution, called the soft octave convolution, which could achieve face generation. Ayala et al. \cite{ayala2019lightweight} realized the recognition of fire scenes by using an octave convolution-based network. Komatsu et al. \cite{komatsu2019octave} used octave convolution to predict depth maps. Gao et al. \cite{gao2020highly} proposed a novel multi-scale octave convolution to achieve high-speed saliency maps generation. The above researches pay more attention to the advantages of the fast speed and the low resource occupancy of octave convolution, and pay more attention to its multi-scale characteristics, ignoring the research on its noise-resilient performance.

\subsection{Noise-resilient and Denoising in Image Processing based on Deep Learning}
\label{sec:Noise-resilientandDenoising}

Unlike images from a video, remote sensing images with noise are usually single-frame imaging, and there is a big difference between two consecutive frames. Therefore, this article mainly discusses single-image processing.
The development of deep neural networks has a significant boost for image denoising algorithms \cite{wang2020practical,zhang2017beyond,zhang2017learning,anwar2019real,cha2019fully,chang2020spatial,tian2020deep}. Most image denoising algorithms regard the denoising process as a process to minimize the empirical risk, given the corrupted inputs $\hat{x}_i$ and the clean targets $y_{i}$, then the process of training this regression model can be formulated as

\begin{equation}
\mathop{\arg\min}_{\theta}\mathop{\Sigma}_{i}L(f_{\theta}(\hat{x}_{i}),y_{i}),
\label{eq:empiricalrisk}
\end{equation}

where $f_{\theta}$ is a parametric family of mappings under the loss function $L$ \cite{lehtinen2018noise2noise}. Therefore, directly applying the denoising algorithms to the tasks of aerial object detection with noise will greatly increase the amount of computational cost. In addition, it is very difficult to obtain enough clean remote sensing images and corresponding noise remote sensing images.

Therefore, how to improve the noise-resilient capability of the object detection network itself is very important \cite{milyaev2017towards}. At present, some works try to solve this problem with the wavelet transform. Aina et al. \cite{ferra2018multiple} and Williams et al. \cite{williams2018wavelet} introduced wavelet transform to the pooling layer. Liu et al. \cite{liu2018multi} designed a multi-level wavelet-CNN by stacking of discrete wavelet transform (DWT), CNNs, and inverse wavelet transform (IWT) modules to achieve noise-resilient capability. For synthetic aperture radar (SAR) images, Duan et al. \cite{duan2017sar} and Gao et al. \cite{gao2019sea} used a combination of a common convolutional layer and a pooling layer based on wavelet transform, which effectively improves the segmentation accuracy of SAR images. Although the combination of wavelet transform and neural networks improves the noise-resilient capability of CNNs, there are still the following problems:

\begin{itemize}
  \item [1)]
  The current changes only exist at the layer level (i.e. pooling layer), and there is a lack of research on the noise-resilient capability of the modules (such as backbone, feature pyramid network, head, etc.). Therefore, it is difficult to guide the newly designed structures.
  \item [2)]
  The explicit transformation between the spatial domain and the frequency domain greatly slows down the training and inference speed of the network.
\end{itemize}

\subsection{Multi-Scale Feature Representations}
\label{sec:Multi-ScaleFeatureRepresentations}

For deep learning-based object detection tasks, the fusion of multi-scale features can effectively improve the detection accuracy of small objects and scale-varying objects. As one of the pioneering works, feature pyramid network (FPN) \cite{lin2017feature} uses a top-down pathway to build a classic, concise pyramid structure. Based on FPN, PANet \cite{liu2018path} adds a new bottom-up feature propagation path, thus forming a structure of two-way interaction between bottom and top features. Following this idea, the adaptively spatial feature fusion (ASFF) \cite{liu1911learning} is designed with a dense multi-scale feature flow method. STDL \cite{zhou2018scale} uses a scale-transfer module to reconstruct feature maps of different scales without increasing the number of parameters. Kong et al. \cite{kong2018deep} first combine features at all scales and reconfigure features by a global attention branch. Similarly, AugFPN \cite{guo2020augfpn} and $U^2-ONet$ \cite{wang2020u2} also integrate feature maps of all scales. Different from the common characteristic pyramid design, NAS-FPN \cite{ghiasi2019fpn} adopts neural architecture search to automatically design the optimal network structure. On the basis of NAS-FPN, MnasFPN \cite{chen2020mnasfpn} optimizes the structure design for mobile devices. The latest BiFPN \cite{tan2020efficientdet} greatly enhances multi-scale feature expression through cross-layer connections and shortcut connections.

However, these studies are still limited in two aspects. (1) They are not designed specifically for noise images. (2) The need for frequency division processing of feature maps is not considered.

\section{Proposed Method}
\label{sec:ProposedMethod}

In this section, we first analyze the noise-resilient theory of neural networks and propose a design theory of noise-resilient structures (Section~\ref{sec:Noise-resilientandDenoising}), and then introduce the main ideas for our proposed OcSaFPN (Section~\ref{sec:Cross-Frequency} and Section~\ref{sec:WeightFeatureFusion}).

\subsection{Noise-resilient Structure Hypotheses}
\label{sec:Noise-resilientStructureTheory}

According to the researches in Section~\ref{sec:Noise-resilientandDenoising}, the frequency division processing of images is feasible in the field of aerial image object detection with noise. From this perspective, CNNs tend to capture the training data in an order from low to high frequencies \cite{xu2019training}. In other words, networks readily fit lower frequencies, but learn higher frequencies later in the training. Besides, since most of the energy in the image is concentrated in the low frequency components, there is a phenomenon that low frequency noise affects the networks more than their high-frequency counterparts \cite{rahaman2019spectral}. Therefore, how to better process the noise data in the low-frequency component will have a huge impact on the performance of the neural network. Furthermore, there are a lot of inherent redundancy in dense model parameters, in the channel dimension of feature maps and in the spatial dimension of feature maps \cite{chen2019drop,han2020ghostnet}. Related to the characteristics of aerial images with noise (Section~\ref{sec:Introduction}), a hypothesis that compresses redundant information in low-frequency components and preserves high-frequency components as much as possible can improve the noise-resilient ability of CNNs is proposed. The explicit wavelet transform-based algorithms in Section~\ref{sec:Noise-resilientandDenoising} can be more time-consuming. In fact, the piecewise nonlinear functions of neural networks can be regarded as implicit wavelet transform. Therefore, we use \href{https://github.com/d-li14/octconv.pytorch}{Oct-ResNet50} as the backbone network, and the experiments in Section~\ref{sec:Experimental} prove the validity of this hypothesis.

Since there is no pyramid structure specifically designed for Oct-ResNet, further analysis of the Octave structure is necessary. Here, the hypothesis that the transmission of the information at different frequency components similar to octave convolution (see Fig.~\ref{fig:octave} for details) is helpful for improving the noise-resilient performance is proposed. The implicit wavelet transform-based algorithms do not guarantee that the feature maps in the Oct-ResNet with average pooling operation strictly follow the distribution of high-frequency components and low-frequency components. Thus, the transmission of the information at different frequency components can achieve a kind of multi-scale learning. In other words, on the one hand, the expression of high-frequency components and low-frequency components can be optimized, and on the other hand, the semantic gap between feature maps of different depths can be better reduced \cite{liu1911learning}. The experiments in Section~\ref{sec:Experimental} prove the validity of this hypothesis.

\begin{figure*}
	\begin{center}
		\includegraphics[width=0.8\linewidth]{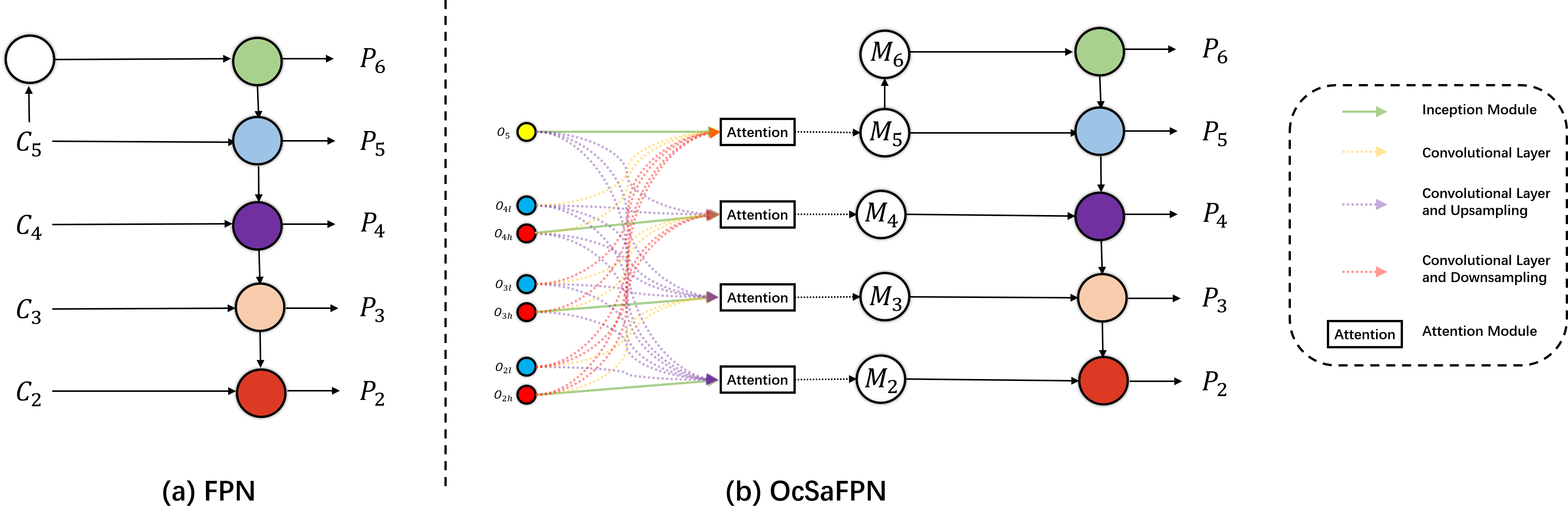}
	\end{center}
	\caption{OcSaFPN:(a) FPN \cite{lin2017feature} uses a top-down pathway to build a classic, concise pyramid structure. (b) The algorithm proposed in this paper, called OcSaFPN, has better noise-resilient capability. }
	\label{fig:OcSaFPN}
\end{figure*}

Based on the above two hypotheses, a novel octave convolution-based semantic attention feature pyramid network (OcSaFPN) is proposed to achieve higher accuracy in aerial images with noise (see Fig.~\ref{fig:OcSaFPN}). Suppose the original image size of the input is $[n,n,3]$, with the input feature maps $\{(O_{2h},O_{2l}),(O_{3h},O_{3l}),(O_{4h},O_{4l}),O_{5}\}$, the feature processing process of OcSaFPN can be expressed as:

%
%
%

\begin{tiny}
\begin{equation}
M_{i}=A\left(CBR\left(Con\left\{
\begin{aligned}
&Inception(O_{j*}), if:i=j \\
&CBR(O_{j**}), if:(i-j)=1 \\
&CUpsample(O_{j**}), if:area(O_{j**}) < area(O_{ih}) \\
&CDownsample(O_{j**}), if:(i-j)>=2
\end{aligned}
\right\}\right)\right)
\label{eq:cross}
\end{equation}
\end{tiny}

\begin{equation}
P_{t}=\left\{
\begin{aligned}
&CBR(Pool(M_{t - 1})), t = 6 \\
&CBR(M_{t}), t = 5 \\
&CBR(M_{t}) \oplus g(M_{t + 1}), t \in [2,5)
\end{aligned}
\right.
\label{eq:fpn}
\end{equation}

where $M_{i}$ denotes the feature map generated by the intermediate process, $P_{t}$ denotes the final feature map from OcSaFPN, $h,l$ denotes high frequency components and low frequency components respectively, $i \in [2,5]$, $j \in [2,5]$, $t \in [2,6]$, $CBR(\cdot)$ denotes the collection of convolution, batch normalization (BN) and ReLu, $Inception(\cdot)$ denotes the inception module \cite{szegedy2015going}, $A(\cdot)$ denotes the attention module \cite{woo2018cbam}, $Con\{\cdot\}$ denotes the concatenation, $CUpsample(\cdot)$ denotes the collection of convolution, BN, ReLu and upsampling operation, $CDownsample(\cdot)$ denotes the collection of convolution, BN, ReLu and downsampling operation, $j*$ means when $i=j=5$, $j*=5$, otherwise, $* \in \{h,l\}$, $j**$ means $** \in \{h,l\}$, and the shapes of $\{(O_{2h},O_{2l}),(O_{3h},O_{3l}),(O_{4h},O_{4l}),O_{5}\}$ are

\begin{equation}\nonumber
\begin{aligned}
&\{([n/4,n/4,128],[n/8,n/8,128]),([n/8,n/8,256], \\
&[n/16,n/16,256]),([n/16,n/16,512],[n/32,n/32,512]), \\
&[n/32,n/32,2048]\}.
\end{aligned}
\end{equation}

In a word, the whole process of OcSaFPN can be described as the information exchange of different frequency components to reduce the semantic gap, and then a new feature fusion strategy is used to realize the construction of the feature pyramid.

\subsection{Cross-Frequency Components Connections}
\label{sec:Cross-Frequency}

To make OcSaFPN seamlessly adapt to the current common two-stage algorithms, the OcSaFPN is designed to receive the feature maps of frequency division as input, and the output feature maps without frequency division. Here, considering that the transmission of information between different frequencies can improve the noise-resilient ability of the structure, a dense connection method is adapted to narrow the semantic gap between feature maps of different depths \cite{liu1911learning}. That means, the high-frequency components and low-frequency components of each convolutional layer are considered as a separate feature map, then each feature map should have an impact on other feature maps, and the weight of this effect will be described in detail in Section~\ref{sec:WeightFeatureFusion}. This kind of connection method can be regarded as a form of redistribution of multi-scale information, and it also conforms to the interaction logic between different frequency components in octave convolution.

\subsection{Weight Feature Fusion}
\label{sec:WeightFeatureFusion}

Since different input features are at different resolutions, they usually contribute to the output feature unequally \cite{tan2020efficientdet}. To address this issue, the characteristics of different channels need to be optimized, and in the spatial dimension, the response degree of different positions also needs to be significant. Based on this, a novel feature fusion strategy is proposed. Our intuition is simple:

\begin{itemize}
  \item [1)]
  The component $O_{i}$ at the same level as the feature map $M_{i}$ should have a greater impact on $M_{i}$. For noisy aerial images, a small amount of detailed information in the high-frequency components should be emphasized, so the Inception module is used to optimize $O_{ih}$.
  \item [2)]
  For the components whose spatial dimension is less than $M_{i}$, we use the collection of convolution, BN, ReLu and bilinear interpolation-based upsampling operation to get new feature maps before concatenation. Here, the collection of convolution, BN, and ReLu is used to reduce the channel dimension and prevent feature degradation. In addition, we chose the bilinear interpolation method in order to make the feature map as smooth as possible, although more detailed information can be obtained by using deconvolutional layer.
  \item [3)]
  Considering the calculation speed, we believe that the difference between $O_{i-1}$ and $M_{i}$ is smaller than that of $O_{i-2}$ and $M_{i}$. Thus, when $(i-1)th$ layer exists, the $O_{(i-1)h}$ and $O_{(i-1)l}$ components can simply be compressed through the CBR module to achieve the compression of the channel dimension and the spectral dimension, and finally transferred to $M_{i}$. For $O_{i}$ and ${M_{t}}$, if t minus i is greater than 1, to reduce the number of convolutional layers, we use the collection of convolution, BN, ReLu , and downsampling operation to achieve the compression of the channel dimension and the spectral dimension.
  \item [4)]
  Feature maps with smaller spatial dimensions should have higher channel dimensions. Thus, before concatenation, we stipulate that the channel dimension of the feature map passed by $O_{i}$ to $\{M_{2},M_{3},M_{4},M_{5}\}$ should increase by a multiple of 2, and the channel dimension of the feature map passed to $M_{2}$ is $32$.
  \item [5)]
  The feature maps after concatenation operation should be sent to attention modules for optimization, and then feature reconstruction is realized through a top-down structure. The addition of non-linear structure can optimize the fused features and prevent the degradation of features.
\end{itemize}

\section{Experimental Results and Analysis}
\label{sec:Experimental}

\subsection{Datasets}
\label{sec:Datasets}
We first evaluate the Oct-ResNet and the proposed OcSaFPN method on DOTA dataset \cite{Xia2018DOTA}. Due to the lack of aerial image datasets with noise, we add different degrees of Gaussian noise and Gaussian blur to the DOTA dataset (additive noise), which degrades the image quality. Finally, we use SSDD dataset \cite{li2017ship} (multiplicative noise) to prove that the proposed OcSaFPN also has better noise-resilient performance on real data.

\textbf{DOTA \cite{Xia2018DOTA}:} This dataset is made up of 2806 large-size aerial images with 15 categories. All images are divided into three datasets including a training set, a validation set, and a test set. The marking method of the sample is a rectangle with a rotation angle. The annotations from the test set are not disclosed, and the inference results can only be evaluated through the online \href{https://captain-whu.github.io/DOTA/evaluation.html}{website}. Due to the limitation of the evaluation times of this website, more experiments will adopt the training set for training and the validation set for accuracy evaluation.

\textbf{DOTANoise:} Different degrees of Gaussian noise and Gaussian blur are added to the DOTA dataset, and a new dataset is finally generated, called DOTANoise.

\textbf{SSDD \cite{li2017ship}:} This dataset is made up of 1160 SAR images with 2456 instances. The size of all images is approximately $500 \times 500$ and the marking form of samples adopts the rectangle without rotation.

\subsection{Evaluation Metrics}
\label{sec:EvaluationMetrics}
For the sake of fairness, we use the standard average precision (AP) value with the intersection over union (IoU) equals to 0.5, which is consistent with the official evaluation methods of the DOTA dataset \cite{Xia2018DOTA} and the SSDD dataset \cite{li2017ship} based on VOC2007 dataset \cite{pascal-voc-2007}. For models trained by DOTA train set and validation set, they are evaluated on the official evaluation \href{https://captain-whu.github.io/DOTA/evaluation.html}{website}. For models trained by DOTA train set and validation set with oriented bounding boxes, they are evaluated on the official evaluation \href{https://captain-whu.github.io/DOTA/evaluation.html}{website}. For models only trained by DOTA train set with oriented bounding boxes, they are evaluated with the official evaluation \href{https://github.com/CAPTAIN-WHU/DOTA_devkit}{tool} on the validation set. For models only trained by SSDD dataset with rectangler bounding boxes, they are evaluated with the \href{https://github.com/open-mmlab/mmdetection}{mmdetection evaluation tool}.

\subsection{Implementation Details}
\label{sec:ImplementationDetails}

For the experiments, we built the baseline network based on Faster R-CNN \cite{Ren2017Faster} and ResNet50 \cite{he2016deep} pretrained on ImageNet \cite{deng2009imagenet}. In order to make the multi-scale features from Oct-ResNet50 be directly used by FPN \cite{2017Feature}, we added a deconvolution layer after each low-frequency component to make it have the same spatial dimension as the corresponding high-frequency component. Besides, BN layers and activation functions are used to suppress the degradation of feature maps after deconvolution layers.

As for DOTA dataset and DOTANoise dataset, original images are cropped into a series of $1024 \times 1024$ patches to reduce the memory requirement. And the images from SSDD dataset, they are resized to $(448,512)$, where 448 represents the length of the short side and 512 the maximum length of an image. To be fair, we use $x,y,\omega,h,\theta$ instead of the two-point coordinate representation of the upper-left and lower-right points in Faster R-CNN for DOTA dataset and DOTANoise dataset.

A series of experiments are designed to better evaluate the efforts of the two hypotheses in Section~\ref{sec:Noise-resilientStructureTheory}, the cross-frequency components connections strategy, the weight feature fusion strategy, and the plug-and-play performance of the OcSaFPN in this paper. The environment is a single NVIDIA Tesla V100 GPU with 16GB memory, along with the PyTorch 1.1.0 and Python 3.7.

\subsection{Ablation Experiments}
\label{sec:AblationExperiments}
Since OcSaFPN is designed follow the hypotheses in Section~\ref{sec:Noise-resilientStructureTheory} from octave convolution, we first demonstrate the noise-resilient performance of Oct-ResNet. Then, in this section, we verify the effectiveness of each component of our proposed network by ablation analysis.

\subsubsection{The noise-resilient performance of octave convolution}
\label{sec:noise-resilientofOct-ResNet}

\begin{table*}[]
\caption{Comparisons with the state-of-the-art detectors in the DOTA dataset \cite{Xia2018DOTA} and DOTA dataset with noise. This table contains a total of 5 groups, and each group uses a different dataset. The $n$ denotes the standard deviation of Gaussian noise. The $v$ denotes the standard deviation of Gaussian blur. In addition, the specific meanings of the following abbreviations are: plane, baseball~ diamond (BD), bridge, ground track field (GTF), small vehicle (SV), large vehicle (LV), ship, tennis court (TC), basketball court (BC), storage tank (ST), soccer-ball field (SBF), roundabout (RA), harbor, swimming pool (SP), and helicopter (HC). All numbers are in $\%$.}
\centering
\resizebox{\textwidth}{!}{
\begin{tabular}{cccccccccccccccccccc}
\toprule
\textbf{Method}& \textbf{Backbone}& \textbf{Dataset}& \textbf{Noise}	& \textbf{Plane}& \textbf{BD}& \textbf{Bridge}& \textbf{GTF}& \textbf{SV}& \textbf{LV}& \textbf{Ship}& \textbf{TC}& \textbf{BC}& \textbf{ST}& \textbf{SBF}& \textbf{RA}& \textbf{Harbor}& \textbf{SP}& \textbf{HC}& \textbf{mAP}\\
\midrule
Faster R-CNN \cite{Ren2017Faster} + FPN	\cite{2017Feature}	& ResNet-50& trainval+test& -& 89.14& 76.35& 49.08& 72.41& 73.81& 72.66& 85.33& 90.85& 79.62& 85.26& 54.6& \textcolor[rgb]{1,0,0}{63.09}& 66.42& 69.53& 60.78& 72.6\\
Faster R-CNN + FPN		& Oct-ResNet-50& trainval+test& -& \textcolor[rgb]{1,0,0}{89.6}& \textcolor[rgb]{1,0,0}{78.72}& 50.16& 67.31& 73.03& 74.81& 85.50& 90.83& 80.66& \textcolor[rgb]{1,0,0}{85.59}& 54.6& 60.74& 66.14& 72.38& 54.84& 72.33\\
Faster R-CNN + OcSaFPN		& Oct-ResNet-50& trainval+test& -& 89.21& 77.06& 49.27& 69& 76.92& 74.33& 85.04& 90.88& 80.77& 85.02& 55.88& 60.26& 67.51& 71.65& 56.04& 72.59\\
RoITransformer \cite{ding2019learning} + FPN		& ResNet-50& trainval+test& -& 88.47& 77.98& \textcolor[rgb]{1,0,0}{54.61}& \textcolor[rgb]{1,0,0}{77.12}& \textcolor[rgb]{1,0,0}{78.18}& \textcolor[rgb]{1,0,0}{77.75}& \textcolor[rgb]{1,0,0}{87.6}& \textcolor[rgb]{1,0,0}{90.89}& \textcolor[rgb]{1,0,0}{86.29}& 85.41& \textcolor[rgb]{1,0,0}{65.86}& 61.24& \textcolor[rgb]{1,0,0}{77.23}& 71.66& \textcolor[rgb]{1,0,0}{62.74}& $\textcolor[rgb]{1,0,0}{76.2}$\\
RoITransformer + OcSaFPN		& Oct-ResNet-50& trainval+test& -& 88.33& 77.06& 52.99& 72.99& 71.51& 76.67& 87.2& 90.87& 79.95& 85.14& 55.98& 61.88& 77.08& \textcolor[rgb]{1,0,0}{72.93}& 56.75& 73.82\\
R3Det \cite{yang2019r3det} + FPN		& ResNet-50& trainval+test& -& \\
R3Det + OcSaFPN		& Oct-ResNet-50& trainval+test& -& \\ \hline

Faster R-CNN + FPN		& ResNet-50& train+val& -& 89.16& 69.06& 37.33& 53.21& 62.29& 74.28& 78.49& 90.8& 60.07& 79.94& 58.88& \textcolor[rgb]{1,0,0}{63.74}& 62.14& 56.75& 53.87& 66\\
Faster R-CNN + FPN		& Oct-ResNet-50& train+val& -& 88.81& 66.5& 40.36& 53.14& 60.49& 73.23& 77.63& 90.46& 59.18& 79.71& 58.01& 62.59& 61.86& 56.76& 48.34& 65.14\\
Faster R-CNN + OcSaFPN		& Oct-ResNet-50& train+val& -& 89.39& 69.13& 39.24& 46.68& 62.4& 75.18& 83.78& 90.53& 64.33& 79.75& 56.64& 59.53& 63.43& 51.35& 49.63& 65.4\\
RoITransformer + FPN		& ResNet-50& train+val& -& 89.93& 68.2& \textcolor[rgb]{1,0,0}{45.54}& \textcolor[rgb]{1,0,0}{70.67}& \textcolor[rgb]{1,0,0}{66.52}& \textcolor[rgb]{1,0,0}{83.5}& \textcolor[rgb]{1,0,0}{87.64}& 90.72& 59.93& \textcolor[rgb]{1,0,0}{85.35}& \textcolor[rgb]{1,0,0}{69.74}& 61.73& \textcolor[rgb]{1,0,0}{75.44}& \textcolor[rgb]{1,0,0}{59.43}& \textcolor[rgb]{1,0,0}{61.38}& \textcolor[rgb]{1,0,0}{71.72}\\
RoITransformer + OcSaFPN		& Oct-ResNet-50& train+val& -& \textcolor[rgb]{1,0,0}{90.00}& \textcolor[rgb]{1,0,0}{71.29}& 43.86& 65.33& 62.29& 77.42& 86.32& \textcolor[rgb]{1,0,0}{90.81}& \textcolor[rgb]{1,0,0}{65.23}& 84.2& 65.89& 63.29& 75.15& 58.78& 49.74& 69.97\\
R3Det + FPN		& ResNet-50& train+val& -& \\
R3Det + OcSaFPN		& Oct-ResNet-50& train+val& -& \\ \hline

Faster R-CNN + FPN		& ResNet-50& train+val& n=0.01,v=1& 79.94& 26.54& 14.44& 11.49& 53.99& 69.55& 69.3& 64.01& 15.51& 61.67& 22.08& 10.93& 48.39& 15.58& 35.71& 39.94\\
Faster R-CNN + FPN		& Oct-ResNet-50& train+val& n=0.01,v=1& 80.08& 26.41& 17.92& 19.89& 51.38& 70.61& 68.68& 68.81& 13.5& 63.3& 19.72& 12.34& 50.15& 15.72& 29.98& 40.57\\
Faster R-CNN + OcSaFPN		& Oct-ResNet-50& train+val& n=0.01,v=1& 80.97& 43.72& 32.31& 27.86& \textcolor[rgb]{1,0,0}{60.35}& 73.91& 76.62& 81.68& 37.16& \textcolor[rgb]{1,0,0}{77.15}& 36.4& 38.28& 59.95& 24.16& \textcolor[rgb]{1,0,0}{47.58}& 53.21\\
RoITransformer + FPN		& ResNet-50& train+val& n=0.01,v=1& 81.17& 47.58& 34.8& 35.9& 58.85& \textcolor[rgb]{1,0,0}{77.25}& \textcolor[rgb]{1,0,0}{79.11}& 81.62& 37.87& 76.3& 37.08& 32.7& 70.77& \textcolor[rgb]{1,0,0}{26.54}& 47.16& 54.98\\
RoITransformer + OcSaFPN		& Oct-ResNet-50& train+val& n=0.01,v=1& \textcolor[rgb]{1,0,0}{88.44}& \textcolor[rgb]{1,0,0}{50.95}& \textcolor[rgb]{1,0,0}{37.98}& \textcolor[rgb]{1,0,0}{37.45}& 58.35& 75.71& 78.08& \textcolor[rgb]{1,0,0}{81.79}& \textcolor[rgb]{1,0,0}{40.9}& 77.1& \textcolor[rgb]{1,0,0}{41.02}& \textcolor[rgb]{1,0,0}{39.55}& \textcolor[rgb]{1,0,0}{72.57}& 23.84& 46.81& \textcolor[rgb]{1,0,0}{56.77}\\
R3Det + FPN		& ResNet-50& train+val& n=0.01,v=1& \\
R3Det + OcSaFPN		& Oct-ResNet-50& train+val& n=0.01,v=1& \\ \hline

Faster R-CNN + FPN		& ResNet-50& train+val& n=0.2,v=0.5& 77.26& 23.55& 9.95& 12.47& 51.42& 67.87& 68.44& 56.01& 9.29& 55.99& 19.15& 9.09& 44.67& 13.97& 26.94& 36.4\\
Faster R-CNN + FPN		& Oct-ResNet-50& train+val& n=0.2,v=0.5& 78.84& 23.98& 12.18& 13.17& 53.38& 68.2& 68.2& 63.45& 10.45& 58.27& 19.53& 9.43& 48.83& 13.89& 31.42& 38.22\\
Faster R-CNN + OcSaFPN		& Oct-ResNet-50& train+val& n=0.2,v=0.5& 80.3& 23.63& 11.21& 11.79& \textcolor[rgb]{1,0,0}{53.57}& 71.78& 69.31& 65.78& 13.69& 58.84& 16.57& \textcolor[rgb]{1,0,0}{10.36}& 51.28& 11.07& 27.09& 38.42\\
RoITransformer + FPN		& ResNet-50& train+val& n=0.2,v=0.5& 79.91& 25.82& 10.27& \textcolor[rgb]{1,0,0}{16.29}& 51.69& 75.7& 70.44& 63.28& 10.71& 56.71& 24.93& 9.73& 61.22& \textcolor[rgb]{1,0,0}{15.79}& 32.47& 40.33\\
RoITransformer + OcSaFPN		& Oct-ResNet-50& train+val& n=0.2,v=0.5& \textcolor[rgb]{1,0,0}{80.92}& \textcolor[rgb]{1,0,0}{26.88}& \textcolor[rgb]{1,0,0}{14.17}& 15.71& 52.83& \textcolor[rgb]{1,0,0}{76.14}& \textcolor[rgb]{1,0,0}{75.68}& \textcolor[rgb]{1,0,0}{69.88}& \textcolor[rgb]{1,0,0}{20.09}& \textcolor[rgb]{1,0,0}{59.79}& \textcolor[rgb]{1,0,0}{31.96}& 9.26& \textcolor[rgb]{1,0,0}{63.13}& 12.09& \textcolor[rgb]{1,0,0}{38.26}& \textcolor[rgb]{1,0,0}{43.12}\\
R3Det + FPN		& ResNet-50& train+val& n=0.2,v=0.5& \\
R3Det + OcSaFPN		& Oct-ResNet-50& train+val& n=0.2,v=0.5& \\ \hline

Faster R-CNN + FPN		& ResNet-50& train+val& n=0.2,v=1& 75.53& 22.51& 9.76& 11.35& 47.65& 68.57& 67.38& 52.57& 7.7& 54.68& 21.59& 9.09& 45.37& 10.59& 24.28& 35.24\\
Faster R-CNN + FPN		& Oct-ResNet-50& train+val& n=0.2,v=1& 78.00& 24.7& 10.49& 11.81& 45.77& 66.91& 66.47& 59.04& 10.67& 55.93& 15.27& 9.72& 45.26& 10.42& 27.31& 35.85\\
Faster R-CNN + OcSaFPN		& Oct-ResNet-50& train+val& n=0.2,v=1& 78.96& 23.79& \textcolor[rgb]{1,0,0}{13.31}& 9.94& \textcolor[rgb]{1,0,0}{50.96}& 71.11& 67.2& 63.57& 16.59& 58.51& 14.6& 9.09& 47.18& 12.22& 31.85& 37.92\\
RoITransformer + FPN		& ResNet-50& train+val& n=0.2,v=1& 79.23& 24.75& 9.95& 14.29& 48.5& 45.34& 69.75& 61.62& 11.20& 57.33& 28.97& 9.45& 60.01& \textcolor[rgb]{1,0,0}{15.1}& 24.67& 39.34\\
RoITransformer + OcSaFPN		& Oct-ResNet-50& train+val& n=0.2,v=1& \textcolor[rgb]{1,0,0}{80.55}& \textcolor[rgb]{1,0,0}{25.61}& 13.2& \textcolor[rgb]{1,0,0}{17.6}& 47.35& \textcolor[rgb]{1,0,0}{74.17}& \textcolor[rgb]{1,0,0}{73.28}& \textcolor[rgb]{1,0,0}{67.26}& \textcolor[rgb]{1,0,0}{17.38}& \textcolor[rgb]{1,0,0}{59.07}& \textcolor[rgb]{1,0,0}{31.96}& \textcolor[rgb]{1,0,0}{10.87}& \textcolor[rgb]{1,0,0}{62.73}& 11.16& \textcolor[rgb]{1,0,0}{32.54}& \textcolor[rgb]{1,0,0}{41.65}\\
R3Det + FPN		& ResNet-50& train+val& n=0.2,v=1& \\
R3Det + OcSaFPN		& Oct-ResNet-50& train+val& n=0.2,v=1& \\ \hline
\bottomrule
\end{tabular}}
\label{tab:results-in-dota}
\footnotesize{Best results are highlighted in $\rm \textcolor[rgb]{1,0,0}{red}$ for every group.}
\end{table*}

To prove the noise-resilient performance of octave convolution, ResNet50 and Oct-ResNet50 are put into the combination of Faster R-CNN and FPN respectively. Here, three deconvolutional layers are added to Oct-ResNet50, which enable the output feature maps of Oct-Resnet50 to enter FPN easily. We use the same class/box prediction network (from Faster R-CNN), the same feature pyramid network, and the same training settings for all experiments. From the first two rows of the first two groups in Table~\ref{tab:results-in-dota}, compared to ResNet50, Oct-Resnet50 shows a reduction in mAP values on noiseless data. However, in the following three groups (the first two rows) with different levels of noise, the mAP values are improved by $0.63\% (n=0.01, v=1)$, $1.82\% (n=0.2, v=0.5)$ and $0.61\% (n=0.2, v=1)$ with Oct-Resnet50 , respectively, compared with Resnet50. Here, the $n$ denotes the standard deviation of Gaussian noise, and the $v$ denotes the standard deviation of Gaussian blur. This proves that the Oct-ResNet50 has better noise-resilient performance than the commonly used ResNet50 structure. At the same time, the method of frequency division and compression of low frequencies is effective to improve the noise-resilient performance of CNNs. This verifies the effectiveness of the hypotheses proposed in Section~\ref{sec:Noise-resilientStructureTheory}, and the accuracy of the network structure (OcSaFPN) designed based on these hypotheses will be evaluated later. Interestingly, we noticed that the accuracy of the categories with fewer samples declined more. For example, the accuracy of the class of Bridge with fewer samples declined faster than the class of Small Vehicle with more samples. This means that simple data enhancement for categories with small sample size may effectively improve the accuracy.

\subsubsection{The effectiveness of the cross-frequency components connections}
\label{sec:noise-resilientofOct-ResNet}

\begin{table*}[]
\caption{Ablation experiments. This table contains a total of four groups, the same data used in each group. For the feature map $M_{i}$, the connections between the feature maps $O_{j}$ and $M_{i}$ are called Near when $j=i \pm 1$, and the connections between the feature maps $O_{j}$ and $M_{i}$ are called Near when $j \neq i \pm 1$. All experiments are based on DOTA dataset and DOTANoise dataset, where the train set is used for training and the validation set is used for evaluation. In addition, the specific meanings of the following abbreviations are: plane, baseball~ diamond (BD), bridge, ground track field (GTF), small vehicle (SV), large vehicle (LV), ship, tennis court (TC), basketball court (BC), storage tank (ST), soccer-ball field (SBF), roundabout (RA), harbor, swimming pool (SP), and helicopter (HC). All numbers are in $\%$.}
\centering
\resizebox{\textwidth}{!}{
\begin{tabular}{ccccccccccccccccccccc}
\toprule
\textbf{Method}& \textbf{Backbone}& \textbf{Near}& \textbf{Far}& \textbf{Noise}	& \textbf{Plane}& \textbf{BD}& \textbf{Bridge}& \textbf{GTF}& \textbf{SV}& \textbf{LV}& \textbf{Ship}& \textbf{TC}& \textbf{BC}& \textbf{ST}& \textbf{SBF}& \textbf{RA}& \textbf{Harbor}& \textbf{SP}& \textbf{HC}& \textbf{mAP}\\
\midrule
Faster R-CNN + OcSaFPN		& Oct-ResNet-50& \checkmark& \checkmark& -& \textcolor[rgb]{1,0,0}{89.39}& 69.13& 39.24& 46.68& \textcolor[rgb]{1,0,0}{62.4}& \textcolor[rgb]{1,0,0}{75.18}& \textcolor[rgb]{1,0,0}{83.78}& 90.53& \textcolor[rgb]{1,0,0}{64.33}& 79.75& \textcolor[rgb]{1,0,0}{56.64}& 59.53& \textcolor[rgb]{1,0,0}{63.43}& 51.35& 49.63& \textcolor[rgb]{1,0,0}{65.4}\\
Faster R-CNN + OcSaFPN		& Oct-ResNet-50& \checkmark& -& -& 88.82& 64.96& 36.96& 49.91& 61.09& 72.38& 76.24& 90.48& 58.77& \textcolor[rgb]{1,0,0}{80.03}& 56.03& \textcolor[rgb]{1,0,0}{65.11}& 59.29& \textcolor[rgb]{1,0,0}{58.01}& \textcolor[rgb]{1,0,0}{50.82}& 64.59\\
Faster R-CNN + OcSaFPN		& Oct-ResNet-50& -& \checkmark& -& 88.68& \textcolor[rgb]{1,0,0}{69.31}& \textcolor[rgb]{1,0,0}{41.86}& \textcolor[rgb]{1,0,0}{59.09}& 62.3& 71.23& 76.47& \textcolor[rgb]{1,0,0}{90.64}& 53.8& 79.55& 48.56& 60.77& 60.53& 57.65& 48.93& 64.62\\ \hline

Faster R-CNN + OcSaFPN		& Oct-ResNet-50& \checkmark& \checkmark& n=0.01,v=1& \textcolor[rgb]{1,0,0}{80.97}& \textcolor[rgb]{1,0,0}{43.72}& \textcolor[rgb]{1,0,0}{32.31}& \textcolor[rgb]{1,0,0}{27.86}& \textcolor[rgb]{1,0,0}{60.35}& \textcolor[rgb]{1,0,0}{73.91}& \textcolor[rgb]{1,0,0}{76.62}& \textcolor[rgb]{1,0,0}{81.68}& \textcolor[rgb]{1,0,0}{37.16}& \textcolor[rgb]{1,0,0}{77.15}& \textcolor[rgb]{1,0,0}{36.4}& \textcolor[rgb]{1,0,0}{38.28}& \textcolor[rgb]{1,0,0}{59.95}& 24.16& 47.58& \textcolor[rgb]{1,0,0}{53.21}\\
Faster R-CNN + OcSaFPN		& Oct-ResNet-50& \checkmark& -& n=0.01,v=1& 80.88& 43.3& 30.74& 22.93& 59.05& 68.79& 76.51& 81.08& 31.42& 74.85& 30.08& 32.49& 58.17& \textcolor[rgb]{1,0,0}{25.14}& \textcolor[rgb]{1,0,0}{49.07}& 50.97\\
Faster R-CNN + OcSaFPN		& Oct-ResNet-50& -& \checkmark& n=0.01,v=1& 80.79& 39.77& 30.23& 24.99& 58.16& 70.86& 75.43& 81.42& 34.33& 76.38& 35.69& 31.73& 57.67& 24.64& 45.77& 51.19\\ \hline

Faster R-CNN + OcSaFPN		& Oct-ResNet-50& \checkmark& \checkmark& n=0.2,v=0.5& \textcolor[rgb]{1,0,0}{80.3}& 23.63& \textcolor[rgb]{1,0,0}{11.21}& \textcolor[rgb]{1,0,0}{11.79}& \textcolor[rgb]{1,0,0}{53.57}& \textcolor[rgb]{1,0,0}{71.78}& \textcolor[rgb]{1,0,0}{69.31}& \textcolor[rgb]{1,0,0}{65.78}& 13.69& 58.84& 16.57& \textcolor[rgb]{1,0,0}{10.36}& \textcolor[rgb]{1,0,0}{51.28}& 11.07& 27.09& \textcolor[rgb]{1,0,0}{38.42}\\
Faster R-CNN + OcSaFPN		& Oct-ResNet-50& \checkmark& -& n=0.2,v=0.5& 78.36& 21.08& 10.24& 9.38& 52.08& 60.56& 66.96& 60.31& \textcolor[rgb]{1,0,0}{13.95}& 58.46& \textcolor[rgb]{1,0,0}{20.38}& 9.49& 46.35& \textcolor[rgb]{1,0,0}{14.01}& 32.69& 36.95\\
Faster R-CNN + OcSaFPN		& Oct-ResNet-50& -& \checkmark& n=0.2,v=0.5& 77.42& \textcolor[rgb]{1,0,0}{26.05}& 10.03& 9.13& 51.65& 63.03& 66.18& 62.65& 13.28& \textcolor[rgb]{1,0,0}{58.95}& 19.47& 9.39& 43.04& 13.25& \textcolor[rgb]{1,0,0}{35.55}& 37.27\\ \hline

Faster R-CNN + OcSaFPN		& Oct-ResNet-50& \checkmark& \checkmark& n=0.2,v=1& \textcolor[rgb]{1,0,0}{78.96}& \textcolor[rgb]{1,0,0}{23.79}& \textcolor[rgb]{1,0,0}{13.31}& 9.94& \textcolor[rgb]{1,0,0}{50.96}& \textcolor[rgb]{1,0,0}{71.11}& \textcolor[rgb]{1,0,0}{67.2}& \textcolor[rgb]{1,0,0}{63.57}& \textcolor[rgb]{1,0,0}{16.59}& \textcolor[rgb]{1,0,0}{58.51}& 14.6& 9.09& \textcolor[rgb]{1,0,0}{47.18}& 12.22& 31.85& \textcolor[rgb]{1,0,0}{37.92}\\
Faster R-CNN + OcSaFPN		& Oct-ResNet-50& \checkmark& -& n=0.2,v=1& 74.17& 23.62& 10.92& \textcolor[rgb]{1,0,0}{13.21}& 45.63& 62.44& 63.39& 61.47& 14.77& 56.57& 18.4& 9.14& 40.03& 11.74& 31.43& 35.79\\
Faster R-CNN + OcSaFPN		& Oct-ResNet-50& -& \checkmark& n=0.2,v=1& 77.07& 23.75& 10.78& 11.09& 45.08& 62.72& 65.51& 58.89& 14.46& 57.3& \textcolor[rgb]{1,0,0}{20.31}& \textcolor[rgb]{1,0,0}{9.17}& 42.71& \textcolor[rgb]{1,0,0}{12.39}& \textcolor[rgb]{1,0,0}{33.17}& 36.29\\ \hline
\bottomrule
\end{tabular}}
\label{tab:far-near}
\footnotesize{Best results are highlighted in $\rm \textcolor[rgb]{1,0,0}{red}$ for every group.}
\end{table*}

Table~\ref{tab:far-near} shows the accuracy for feature networks with different cross-frequency components connections. As shown in Fig.~\ref{fig:OcSaFPN}(b), OcSaFPN adopts a dense connect strategy to realize the interaction of different frequency feature maps from $O_{j}$ to $M_{i}$. To analysis the efficiency of the cross-frequency components connections, these connections are divided into two groups. For the feature map $M_{i}$, the connections between the feature maps $O_{j}$ and $M_{i}$ are called Near when $j=i \pm 1$, and the connections between the feature maps $O_{j}$ and $M_{i}$ are called Near when $j\neq i \pm 1$. For all the experiments, there is only the difference in OcSaFPN. Here, the DOTA training set is used for training and the DOTA validation set is used for accuracy evaluation. Based on this configuration, OcSaFPN with both Far and Near connections achieves the highest mAP values under different noise conditions. This means that the cross-frequency components connections of different depths are valid. Besides, compared to the second-highest mAP value, the mAP value of OcSaFPN with both Far and Near connections improves only $0.78\%$ with noiseless images. But on noisy images, the mAP value of OcSaFPN with both Far and Near connections improves $1.15\% (n=0.2,v=0.5)$, $1.63\% (n=0.2,v=1)$, and $2.02\% (n=0.01,v=1)$. Thus, it's important to consider the Near and Far connections at the same time for aerial object detection tasks with noise. In addition, OcSaFPN with only Far connections obtains three second-highest mAP values under four different noise conditions, and OcSaFPN with only Near connections obtains only one second-highest mAP values under four different noise conditions. This means that features from different depths have different influences and need to be treated differently. For specific categories of target detection tasks, the trade-off between Near and Far can be used to strike a balance between computational effort and accuracy.

\subsubsection{The effectiveness of the weight feature fusion strategy}
\label{sec:effectivenessweightfeaturefusion}

\begin{table*}[]
\caption{Ablation experiments. This table contains a total of four groups, the same data used in each group. The $n$ denotes the standard deviation of Gaussian noise. The $v$ denotes the standard deviation of Gaussian blur. The $Attention$ denotes whether to use the attention module in Eq~\ref{eq:cross}, and the $Fusion$ denotes whether to use the fusion strategy in Section~\ref{sec:WeightFeatureFusion}. All experiments are based on DOTA dataset and DOTANoise dataset, where the train set is used for training and the validation set is used for evaluation. In addition, the specific meanings of the following abbreviations are: plane, baseball~ diamond (BD), bridge, ground track field (GTF), small vehicle (SV), large vehicle (LV), ship, tennis court (TC), basketball court (BC), storage tank (ST), soccer-ball field (SBF), roundabout (RA), harbor, swimming pool (SP), and helicopter (HC). All numbers are in $\%$.}
\centering
\resizebox{\textwidth}{!}{
\begin{tabular}{ccccccccccccccccccccc}
\toprule
\textbf{Method}& \textbf{Backbone}& \textbf{Attention}& \textbf{Fusion}& \textbf{Noise}	& \textbf{Plane}& \textbf{BD}& \textbf{Bridge}& \textbf{GTF}& \textbf{SV}& \textbf{LV}& \textbf{Ship}& \textbf{TC}& \textbf{BC}& \textbf{ST}& \textbf{SBF}& \textbf{RA}& \textbf{Harbor}& \textbf{SP}& \textbf{HC}& \textbf{mAP}\\
\midrule
Faster R-CNN + OcSaFPN		& Oct-ResNet-50& \checkmark& \checkmark& -& \textcolor[rgb]{1,0,0}{89.39}& \textcolor[rgb]{1,0,0}{69.13}& \textcolor[rgb]{1,0,0}{39.24}& 46.68& 62.4& \textcolor[rgb]{1,0,0}{75.18}& \textcolor[rgb]{1,0,0}{83.78}& 90.53& \textcolor[rgb]{1,0,0}{64.33}& 79.75& 56.64& 59.53& \textcolor[rgb]{1,0,0}{63.43}& 51.35& 49.63& 65.4\\
Faster R-CNN + OcSaFPN		& Oct-ResNet-50& \checkmark& -& -& 88.54& 66.07& 33.7& \textcolor[rgb]{1,0,0}{54.93}& 60.31& 72.66& 76.35& \textcolor[rgb]{1,0,0}{90.64}& 57.86& \textcolor[rgb]{1,0,0}{84.77}& \textcolor[rgb]{1,0,0}{62.92}& 60.75& 60.23& \textcolor[rgb]{1,0,0}{59.44}& 49.82& 65.27\\
Faster R-CNN + OcSaFPN		& Oct-ResNet-50& -& \checkmark& -& 88.48& 65.72& 38.25& 54.43& \textcolor[rgb]{1,0,0}{62.69}& 72.49& 76.83& 90.41& 61.72& 79.36& 57.96& \textcolor[rgb]{1,0,0}{66.14}& 60.42& 58.57& \textcolor[rgb]{1,0,0}{55.95}& \textcolor[rgb]{1,0,0}{65.96}\\ \hline

Faster R-CNN + OcSaFPN		& Oct-ResNet-50& \checkmark& \checkmark& n=0.01,v=1& \textcolor[rgb]{1,0,0}{80.97}& 43.72& 32.31& \textcolor[rgb]{1,0,0}{27.86}& \textcolor[rgb]{1,0,0}{60.35}& \textcolor[rgb]{1,0,0}{73.91}& 76.62& \textcolor[rgb]{1,0,0}{81.68}& \textcolor[rgb]{1,0,0}{37.16}& \textcolor[rgb]{1,0,0}{77.15}& \textcolor[rgb]{1,0,0}{36.4}& \textcolor[rgb]{1,0,0}{38.28}& \textcolor[rgb]{1,0,0}{59.95}& 24.16& 47.58& \textcolor[rgb]{1,0,0}{53.21}\\
Faster R-CNN + OcSaFPN		& Oct-ResNet-50& \checkmark& -& n=0.01,v=1& 80.15& \textcolor[rgb]{1,0,0}{43.93}& \textcolor[rgb]{1,0,0}{32.63}& 22.27& 56.08& 69.59& 76.61& 81.12& 36.97& 75.34& 28.7& 33.07& 58.91& 25.11& 47.28& 51.18\\
Faster R-CNN + OcSaFPN		& Oct-ResNet-50& -& \checkmark& n=0.01,v=1& 80.44& 40.83& 31.37& 25.01& 55.62& 69.23& \textcolor[rgb]{1,0,0}{76.68}& 81.18& 34.7& 74.53& 26.11& 30.7& 58.77& \textcolor[rgb]{1,0,0}{25.83}& \textcolor[rgb]{1,0,0}{51}& 50.82\\ \hline

Faster R-CNN + OcSaFPN		& Oct-ResNet-50& \checkmark& \checkmark& n=0.2,v=0.5& \textcolor[rgb]{1,0,0}{80.3}& 23.63& \textcolor[rgb]{1,0,0}{11.21}& 11.79& \textcolor[rgb]{1,0,0}{53.57}& \textcolor[rgb]{1,0,0}{71.78}& \textcolor[rgb]{1,0,0}{69.31}& \textcolor[rgb]{1,0,0}{65.78}& \textcolor[rgb]{1,0,0}{13.69}& \textcolor[rgb]{1,0,0}{58.84}& 16.57& \textcolor[rgb]{1,0,0}{10.36}& \textcolor[rgb]{1,0,0}{51.28}& 11.07& 27.09& \textcolor[rgb]{1,0,0}{38.42}\\
Faster R-CNN + OcSaFPN		& Oct-ResNet-50& \checkmark& -& n=0.2,v=0.5& 78.52& \textcolor[rgb]{1,0,0}{24.51}& 8.41& \textcolor[rgb]{1,0,0}{14.77}& 47.61& 63.22& 68.09& 61.18& 10.66& 56.13& \textcolor[rgb]{1,0,0}{19.01}& 9.5& 47.43& 8.9& \textcolor[rgb]{1,0,0}{36.23}& 36.95\\
Faster R-CNN + OcSaFPN		& Oct-ResNet-50& -& \checkmark& n=0.2,v=0.5& 76.96& 22.45& 10.28& 11.81& 50.24& 62.83& 67.88& 59.93& 10.38& 57.15& 17.41& 9.36& 46.07& \textcolor[rgb]{1,0,0}{12.78}& 34.88& 36.69\\ \hline

Faster R-CNN + OcSaFPN		& Oct-ResNet-50& \checkmark& \checkmark& n=0.2,v=1& \textcolor[rgb]{1,0,0}{78.96}& \textcolor[rgb]{1,0,0}{23.79}& \textcolor[rgb]{1,0,0}{13.31}& 9.94& \textcolor[rgb]{1,0,0}{50.96}& \textcolor[rgb]{1,0,0}{71.11}& \textcolor[rgb]{1,0,0}{67.2}& \textcolor[rgb]{1,0,0}{63.57}& \textcolor[rgb]{1,0,0}{16.59}& \textcolor[rgb]{1,0,0}{58.51}& 14.6& 9.09& \textcolor[rgb]{1,0,0}{47.18}& 12.22& \textcolor[rgb]{1,0,0}{31.85}& \textcolor[rgb]{1,0,0}{37.92}\\
Faster R-CNN + OcSaFPN		& Oct-ResNet-50& \checkmark& -& n=0.2,v=1& 76.88& 21.23& 10.61& \textcolor[rgb]{1,0,0}{11.74}& 47.31& 62.86& 65.66& 58.83& 10.73& 53.58& \textcolor[rgb]{1,0,0}{16.19}& 9.29& 40.88& 12.19& 25.84& 34.92\\
Faster R-CNN + OcSaFPN		& Oct-ResNet-50& -& \checkmark& n=0.2,v=1& 76.69& 22.26& 10.67& 11.22& 46.36& 63.78& 65.43& 59.05& 13.38& 54.74& 14.15& \textcolor[rgb]{1,0,0}{9.39}& 42.46& \textcolor[rgb]{1,0,0}{12.66}& 26& 35.22\\ \hline
\bottomrule
\end{tabular}}
\label{tab:cross-attention}
\footnotesize{Best results are highlighted in $\rm \textcolor[rgb]{1,0,0}{red}$ for every group.}
\end{table*}

As discussed in Section~\ref{sec:WeightFeatureFusion}, a weight feature fusion strategy (Eq~\ref{eq:cross}) is proposed to treat the features from different depths differently. The Table~\ref{tab:cross-attention} compares the effective of the attention module and the fusion strategy. Where the Attention denotes whether to use the attention module in Eq~\ref{eq:cross}, and the Fusion denotes whether to treat the features from different depths differently. As shown in the results, OcSaFPN with both Attention and Fusion achieves three highest mAP values in four different noise conditions. Without Fusion, the mAP value is reduced by $0.13\%$ on clean images, and the mAP values are reduced by $2.03\%$, $1.47\%$, and $3\%$ on different levels noise images. The Fusion can significantly improve the anti-noise performance of the network. Interestingly enough, the OcSaFPN without Attention get a highest mAP value ($65.96\%$) on noiseless images, but its mAP values reduce significantly on noise images. Thus, the attention module can help keep the network stable in different noisy conditions.

\subsection{OcSaFPN for Object Detection with Noise}
\label{sec:OcSaFPNObjectDetectionNoise}

Table~\ref{tab:results-in-dota} shows the experimental results obtained with the DOTA dataset, all parameters remain the same. Starting from a Faster R-CNN detector \cite{Ren2017Faster} with the top-down FPN \cite{2017Feature}, the Oct-ResNet50 shows its superiority in object detection tasks with noise disscussed in Section~\ref{sec:noise-resilientofOct-ResNet}. By further replacing FPN with our proposed OcSaFPN, the mAP values achieve additional $12.64\% (n=0.01, v=1)$, $0.2\% (n=0.2, v=0.5)$, and $2.07\% (n=0.2, v=1)$, where the $n$ denotes the standard deviation of Gaussian noise, and the $v$ denotes the standard deviation of Gaussian blur. Besides, the mAP values of the Faster R-CNN with Oct-ResNet50 and OcSaFPN are $0.26\%$ and $0.26\%$ higher than the mAP values of the Faster R-CNN with Oct-ResNet50 and FPN in DOTA validation set and DOTA test set without noise. Although the mAP values of the Faster R-CNN with Oct-ResNet50 and OcSaFPN are still $0.6\%$ (DOTA validation set) and $0.01\%$ (DOTA test set) lower than the mAP values of the Faster R-CNN with ResNet50 and FPN, considering that OcSaFPN is designed for object detection tasks with noise, this values gap is actually not that large. In addition, the mAP values of the Faster R-CNN with Oct-ResNet50 and OcSaFPN are $13.27\% (n=0.01, v=1)$, $2.02\% (n=0.2, v=0.5)$ and $2.68\% (n=0.2, v=1)$ higher than the mAP values of the Faster R-CNN with ResNet50 and FPN in DOTANoise validation set.

\begin{figure*}
	\begin{center}
		\includegraphics[width=0.8\linewidth]{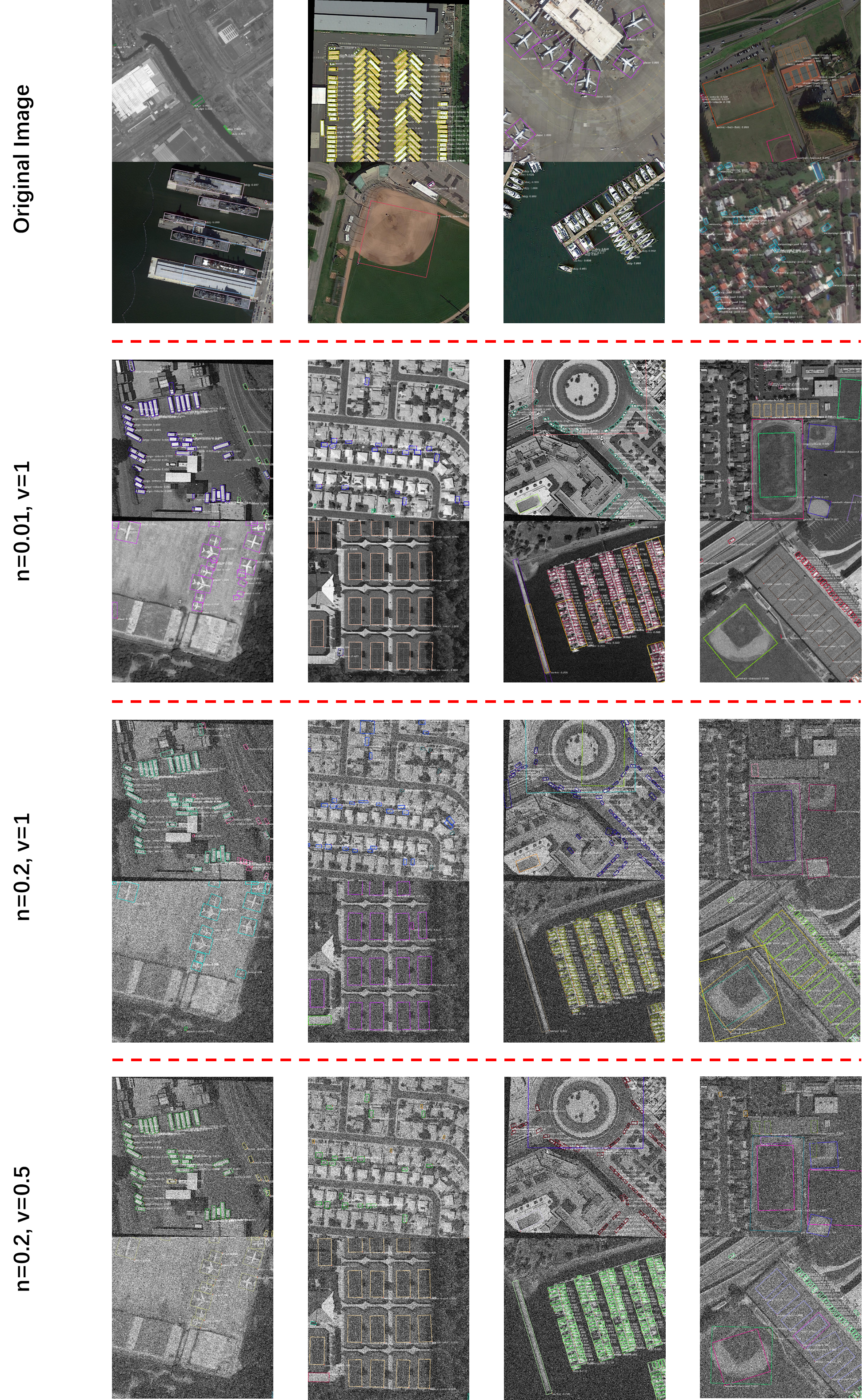}
	\end{center}
	\caption{Visualization of the detection results in the DOTA dataset \cite{Xia2018DOTA} with different noise conditions, where the $n$ denotes the standard deviation of Gaussian noise and the $v$ denotes the standard deviation of Gaussian blur. The method used is the RoITransformer \cite{ding2019learning} with Oct-ResNet50 \cite{chen2019drop} and OcSaFPN.}
	\label{fig:dota_inference}
\end{figure*}

To prove the practicability of the Oct-ResNet and OcSaFPN noise-resilient combination proposed in this paper, we put this combination into two state-of-the-art detectors, RoITransformer\cite{ding2019learning} and R3Det\cite{yang2019r3det}. The mAP values from the last four rows of each group in Table~\ref{tab:results-in-dota} prove the efficiency of the combination. Fig.~\ref{fig:dota_inference} shows the detection results of the combination.

\begin{table}[h]	
	\centering	
	\caption{ Results on SSDD \cite{li2017ship}. All methods are based on Faster R-CNN\cite{Ren2017Faster}. All numbers are in $\%$.}
	\label{tab:SSDD}
	\begin{tabular}{ccccc}
		\toprule
		\multirow{1}{*}{} & \multicolumn{4}{c}{Ship Detection in SAR Dataset} \\
		\cmidrule(r){3-5}
		&  Backbone     &  $AP$   &   Recall & Precision \\
		\midrule
		FPN             &ResNet50   &90.5   &92   &80.9 \\
        FPN             &Oct-ResNet50   &88.1   &89.2   &$\textcolor[rgb]{1,0,0}{87}$ \\
		OcSaFPN (Ours)             &Oct-ResNet50   &$\textcolor[rgb]{1,0,0}{91.8}$   &$\textcolor[rgb]{1,0,0}{93.3}$ &85.3 \\
		\bottomrule
	\end{tabular}
	\hspace{4cm}
	\footnotesize{
		\\ Best results are highlighted in $\rm \textcolor[rgb]{1,0,0}{red}$.}
\end{table}

\begin{figure*}
	\begin{center}
		\includegraphics[width=0.8\linewidth]{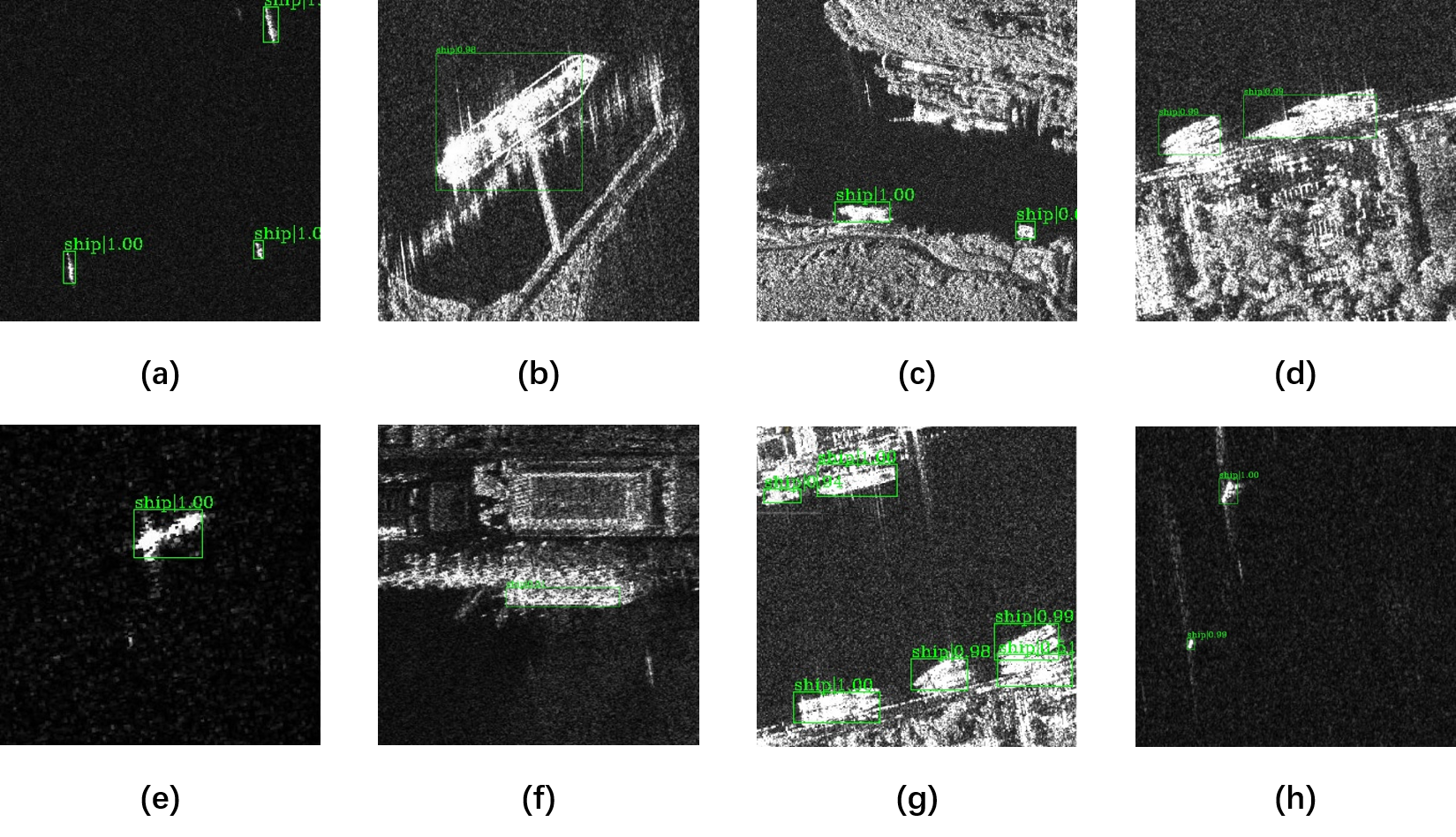}
	\end{center}
	\caption{Visualization of the detection results in the SSDD dataset \cite{li2017ship}. The method used is the Faster R-CNN \cite{Ren2017Faster} with Oct-ResNet50 \cite{chen2019drop} and OcSaFPN.}
	\label{fig:ssdd_inference}
\end{figure*}

Since the above experiments are conducted on a synthetic noise dataset, SAR images containing multiplicative noise are selected to demonstrate the effectiveness of the proposed method on real aerial images. As can be observed in Table~\ref{tab:SSDD}, the combination of Oct-ResNet50 and OcSaFPN achieves highest values in mAP, recall and precision. Fig.~\ref{fig:ssdd_inference} shows the detection results of the combination with Faster R-CNN.

\section{Conclusion}
\label{sec:Conclusion}
In this paper, we analysis the noise-resilient performance of Oct-ResNet, and present a novel octave convolution-based semantic attention feature pyramid network (OcSaFPN), in order to improve the accuracy of aerial object detection tasks with noise. The experiments undertaken with the DOTA dataset, DOTANoise dataset, and SSDD confirmed the remarkable performance of the proposed method on noise aerial images. More experiments have also demonstrated the ease of use of the Oct-ResNet and OcSaFPN combination. In addition, how to simplify OcSaFPN structure and propose a more efficient structure remain to be studied.

\section*{Acknowledgment}

The numerical calculations in this paper have been done on the supercomputing system in the Supercomputing Center of Wuhan University.

\ifCLASSOPTIONcaptionsoff
  \newpage
\fi



%
%
%

\bibliographystyle{IEEEtran}
\bibliography{IEEEabrv,egbib}

%

%
%
%




\end{document}